\documentclass[11pt]{article}

\usepackage[margin=1in]{geometry}
\usepackage[T1]{fontenc}
\usepackage{lmodern}
\usepackage{microtype}

\usepackage{amsmath}
\usepackage{amssymb}
\usepackage{amsthm}
\usepackage{graphicx}
\usepackage{booktabs}
\usepackage{tabularx}

\usepackage[section]{placeins}

\usepackage{tikz}
\usetikzlibrary{positioning,shapes,arrows,arrows.meta,calc}

\usepackage{comment}
\usepackage{listings}
\usepackage{xcolor}
\usepackage[numbers,sort&compress]{natbib}
\usepackage{xurl}
\usepackage{authblk}
\usepackage[hidelinks]{hyperref}

\graphicspath{{Fig/}}
\lstdefinestyle{promptstyle}{
    basicstyle=\ttfamily\tiny,
    breaklines=true,
    breakatwhitespace=false,
    columns=fullflexible,
    keepspaces=true,
    showstringspaces=false,
    frame=single,
    framesep=2pt,
    xleftmargin=0pt,
    xrightmargin=0pt,
    aboveskip=4pt,
    belowskip=4pt,
    linewidth=\columnwidth
}

\theoremstyle{plain}

\theoremstyle{definition}

\theoremstyle{remark}

\title{Reproducibility is not construct validity:\\ LLM measurement of institutionally situated communication}

\author[1]{Veronika Batzdorfer\thanks{Corresponding author: \href{mailto:batzdorfer.veronika@gmail.com}{batzdorfer.veronika@gmail.com}}}
\author[2,3]{Carlo R. M. A. Santagiustina}
\affil[1]{Department of Sociology and Computational Sociology, Karlsruhe Institute of Technology (KIT), Douglasstr. 24, 76133 Karlsruhe, Germany}
\affil[2]{Inria Paris Centre, Inria, 48 rue Barrault, 75013, Paris, France}
\affil[3]{médialab, Sciences Po, 1 Saint-Thomas, 75007,Paris, France}
\date{}

\begin{document}
\maketitle

\begin{abstract}
High annotation reproducibility does not necessarily imply that an LLM-inferred measure captures the construct it is intended to measure. We test this distinction using a dataset from the European Commission's AI Act consultation, linking structured survey responses to free-text consultation submissions from the same stakeholders. LLM annotations of consultation submissions are highly reproducible ($\text{intraclass correlations} > 0.99$), yet show limited convergence with survey-reported measures of the nominal construct they were intended to approximate. Divergence between survey- and LLM-inferred text-based measures varies systematically across stakeholder groups: business associations express greater concern about AI risks in text-based consultations than in survey responses ($\bar{g} = +1.0$), whereas public authorities and several non-business groups show smaller or negative divergences. Divergences between scores suggest positive spatial autocorrelation across European countries (Moran's $I = 0.347$, $p = 0.036$), indicating that stakeholders from neighboring countries tend toward more similar text-based stances towards AI safety concerns. Despite divergence, survey-reported concerns remain strongly associated with support for explainability across all divergence levels. These results demonstrate that LLM annotation reproducibility can coexist with poor construct correspondence and motivate validation procedures that distinguish reproducibility, construct validity, and communication context variation when LLMs are used as measurement instruments.
\end{abstract}


\medskip
\noindent\textbf{Implications for Research and Policy.} What policy stakeholders say publicly may differ from what they report in surveys, posing a challenge for inferring social and political constructs from web-based text. We link public consultation submissions and survey responses from the same stakeholders in the European Commission’s AI Act consultation. The sources diverge systematically across institutional groups: business associations and organizations emphasize AI risks more strongly in public consultations than in surveys, while other groups show smaller or reversed differences. These findings suggest that LLM-based text measures capture not only underlying constructs but also institutional roles and communicative contexts. Rather than treating divergence between text- and survey-based measures as measurement failure, we show how it reveals how stakeholders express and frame concerns differently across communicative settings.

\bigskip
\section{Introduction}

Large language models (LLMs) are increasingly used to convert unstructured text into quantitative measures of political attitudes, preferences, risk concerns, and other latent social constructs ~\cite{baumann2025large, benoit2025using}. Their attractiveness as measurement instruments is partly methodological: unlike conventional hand coding, LLM-based annotation can be applied and scaled rapidly and can produce highly reproducible scores. Yet reproducibility and construct validity are distinct properties~\cite{lin2026validity}. An annotation procedure may generate nearly identical judgments across repeated runs while systematically measuring a feature of the text other than the construct researchers intend to recover.
This distinction is particularly important when text is public and produced in institutional settings. Regulatory consultations, parliamentary submissions, corporate position papers, and public comments are not neutral containers of individual attitudes and preferences ~\cite{bunea2024understanding}. They are communicative artifacts produced for particular audiences, under institutional conventions and with incentives that differ from those governing standardized survey responses~\cite{kuran1995private}. Consequently, a textual measure may capture not only substantive preferences but also issue selection and framing, rhetorical notions, genre conventions, organizational roles, and expressions tailored to recipients~\cite{bunea2014explaining, blair2020worry}.\\

We examine this problem using free-text consultation submissions that were linked to structured survey responses from the same stakeholders participating in the European Commission's consultation on the proposed Artificial Intelligence Act. This linkage allows us to ask whether an LLM annotation of a nominally corresponding textual construct converges with a standardized survey measure reported by the same stakeholder. The design therefore provides a direct test of a central assumption in text-as-data research: that a reproducible LLM score can serve as a proxy for an underlying social construct when the construct is expressed through institutionally situated language.\\

Our research questions lie at the intersection of political communication, the elicitation and measurement of AI-risk concerns, and LLM-assisted computational social science. We ask: \\
\textbf{RQ1}: How closely do LLM-derived measures from consultation texts correspond to survey-reported measures of AI safety, rights, and explainability concerns?  \\
\textbf{RQ2}: Are misalignments between textual and survey measures systematically patterned across institutional and geographic contexts?\\

We find a pronounced reproducibility--validity dissociation.
Across five independent annotation runs, LLM scores are highly reproducible, with intraclass correlations exceeding $0.994$. However, convergence with the corresponding survey measures is weak: correlations range from $0.029$ to $0.176$, while Lin's concordance coefficients remain below $0.08$. The cross-setting divergence is substantively interpretable because the two measures draw on different manifestations of the same nominal construct: the annotation prompt captures the presence and rhetorical salience of concern in public text, whereas the survey elicits its reported intensity. The two procedures may therefore be highly reproducible while targeting different aspects of communication and hence remaining only partially equivalent in what they measure.

The text--survey divergence is also systematically patterned. Business associations exhibit the largest positive divergence (i.e., greater public-facing amplification of AI related concerns relative to survey-reported positions), whereas public authorities and several non-business groups exhibit smaller or negative misalignment. We treat these differences as evidence of institutionally patterned public expression rather than as direct measures of strategic intent. Geographic analyses provide additional descriptive evidence of clustering, although the country-level result is sensitive to multiple-testing correction and should therefore be interpreted cautiously.
\paragraph{Contributions}
We show why LLM annotations reproducibility cannot substitute construct validation when textual measures are used as proxies for social constructs. A LLM-derived text-based measure can remain informative even when it differs from survey responses, because the cross-setting divergence can reveal how stakeholders communicate differently in standardized surveys and public institutional settings
~\cite{jensen2015rhetoric,bunea2014explaining}. The appropriate interpretation, however, depends on the measurement target. Researchers seeking to recover private or survey-reported attitudes should seek to validate text-derived measures against an external criterion; researchers interested in public communication may instead treat the LLM-derived text-based measure as an indicator of situated expressions of the targeted construct. Confusing these objectives risks attributing construct validity to LLM-derived variables beyond what the annotation procedure establishes.

\section{Related Work and Conceptual Framework}

\subsection{LLMs as measurement instruments}

LLMs are increasingly used to label higher-order constructs in text such as attitudes and opinions~\cite{benoit2025using}, with zero-shot performance often exceeding crowd workers~\cite{gilardi2023chatgpt}. Yet validation typically stops at inter-coder agreement~\cite{le2025positioning}, treating label alignment and measurement correlation with human majority votes, or average expert judgment, as sufficient. What remains underexamined is whether the LLM and the survey are measuring the same psychometric construct~\cite{lin2025prompts}---a distinction that becomes critical when the label enters analysis rather than remaining a descriptive tag~\cite{wooddoughty2018challenges, bean2026measuring}. 
But as pointed out by Baden et al.~\cite{baden2022three}, strong performance in matching human annotations or given ground truths does not establish substantive validity, and recent measurement frameworks~\cite{lin2026validity} argue that employing LLMs counts as measurement only when validation can show that it tracks the intended concept rather than the model's priors.\\

Using text-derived variables raises distinct challenges of high dimensionality, latent confounding, and unobserved heterogeneity~\cite{egami2018make,veitch2020adapting}. Prior work has developed text-based adjustment through matching~\cite{roberts2020adjusting}, causally sufficient embeddings~\cite{veitch2020adapting}, and review frameworks for text-as-confounder designs~\cite{keith2020text}. However, this literature often assumes that once a text-derived variable is constructed, it is measured without systematic error relative to the intended latent construct. Whereas, in public facing communicative settings, such as regulatory consultations, the LLM-derived proxy may also capture public rhetoric, institutional positioning and strategic signaling, while the survey will elicit attitudes through a standardized (non public facing) setting. and the gap between them may reflect systematic misalignment structured by stakeholder type~\cite{bunea2014explaining} or country~\cite{austgulen2020understanding}, rather than random measurement error.

\subsection{From annotation reproducibility to construct validity}

\begin{figure}[t]
\centering
\begin{tikzpicture}[
    >=Stealth,
    latent/.style={
        circle,
        draw,
        thick,
        dashed,
        fill=gray!15,
        minimum size=0.9cm,
        inner sep=0pt,
        font=\small\bfseries
    },
    observed/.style={
        circle,
        draw,
        thick,
        fill=white,
        minimum size=0.9cm,
        inner sep=0pt,
        font=\small\bfseries
    },
    proxy/.style={
        rectangle,
        rounded corners=3pt,
        draw,
        thick,
        fill=blue!5,
        minimum width=1.1cm,
        minimum height=0.7cm,
        inner sep=2pt,
        font=\small\bfseries
    },
    arrow/.style={
        ->,
        thick,
        shorten >=1.5pt,
        shorten <=1.5pt,
        black!70
    }
]

    
    \node[latent] (U) at (0, 2.2) {$U$};
    
    \node[latent] (T) at (-1.5, 0.9) {$T$};
    \node[latent] (Y) at (1.5, 0.9) {$Y$};
    
    \node[observed] (X) at (-2.8, -0.3) {$X$};

    \node[observed] (Text) at (0, -0.4) {Text};
    
    \node[proxy] (WT) at (-1.0, -1.8) {$W_T$};
    \node[proxy] (WY) at (1.0, -1.8) {$W_Y$};

    
    \draw[arrow] (U) -- (T);
    \draw[arrow] (U) -- (Y);
    \draw[arrow] (U) -- (Text);
    
    \draw[arrow] (T) -- (Y);
    
    \draw[arrow] (X) -- (T);
    \draw[arrow] (X) -- (Y);
    
    \draw[arrow] (T) -- (Text);
    \draw[arrow] (Y) -- (Text);
    
    \draw[arrow] (Text) -- (WT);
    \draw[arrow] (Text) -- (WY);

    \node[above=0.05cm of U, font=\tiny, align=center] {Unobserved\\confounding};
    \node[left=0.05cm of X, font=\tiny, align=right] {Observed\\covariates};

\end{tikzpicture}
\caption{
DAG for text-derived measurement. $T$ denotes latent AI safety concern and $Y$ the latent AI explainability concern; $U$ denotes unobserved confounding; $X$ observed covariates; Text denotes a public consultation submission; and $W_T, W_Y$ denote LLM annotations. Because Text is a downstream common effect of $T$ and $Y$ ($T \rightarrow \mathrm{Text} \leftarrow Y$), conditioning on text-derived features can induce collider bias, creating bias beyond that arising from measurement error alone.}
\label{fig:dag}
\end{figure}
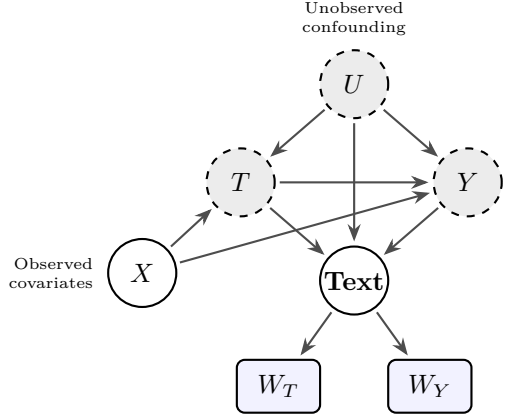

To formalize why LLM-derived text scores may diverge from survey-reported attitudes  towards AI, consider the directed acyclic graph (DAG) presented in Fig.~\ref{fig:dag}. In standard text-as-data research designs, scholars implicitly treat text-derived proxies ($W_T$ and $W_Y$) as direct reflections of stakeholder's latent attitudes ($T$ and $Y$). However, consultation texts are downstream artifacts jointly caused by the stakeholder's latent AI safety ($T$) and explainability ($Y$) concerns, making the document text a structural collider ($T \rightarrow \text{Text} \leftarrow Y$).

Because the LLM proxies ($W_T$ and $W_Y$) are derived from public text, they reflect how the underlying constructs are expressed in specific public discourse contexts. Consequently, these measures may capture not only the underlying construct ($T$), but also noise and systematic biases introduced by the structure of public discourse and related sociocultural factors. Therefore, treating LLM-derived text scores as error-free proxies for the underlying constructs, measurement instruments or controls in downstream models can introduce collider-stratification bias, opening spurious backdoor paths between explanatory variables and outcomes. High inter-run agreement does not resolve this problem, as the reproducibility of LLM-derived text annotations may not translate into construct validity against survey instruments: the model may reliably scale the public expression, without necessarily recovering the attitudinal and psychological construct that the survey is designed to elicit. Moreover, survey responses themselves are shaped by question wording, format, and context~\cite{schwarz1999self}.

\subsection{Public and institutionally situated communication}

It has been shown~\cite{beyers2004voice, bunea2014explaining} that interest-groups and policy stakeholders often tailor their communication strategies to the expectations of relevant audiences (i.e., those who can affect or are affected by them). Recent public administration scholarship shows that reputation management is inherently strategic, with organizations attempting to ``guide the crowd'' rather than merely react to external judgments~\cite{bach2022regulatory, binderkrantz2024core}. Research on interest-group political communication also shows that organizations differ in how they publicly articulate their policy preferences and positions based on organizational incentives and institutional opportunities~\cite{beyers2004voice,dur2013gaining},  but also socioculturally situated repertoires of interpretation and justification ~\cite{swidler1986culture}; through which stakeholders' policy positions and concerns are framed and expressed in the public sphere. We extend this logic to regulatory consultations, proposing that misalignment between survey-reported and publicly expressed concerns may arise through three mechanisms detailed below.

\paragraph{Multiple Audiences}

Public consultation submissions are communicated in a multi-audience environment: regulators evaluate policy arguments, competitors monitor positioning, while customers, supporters and investors assess their alignment to expressed values and objectives~\cite{binderkrantz2024core}. Within this environment, organizational ties can facilitate coordination and public position alignment among stakeholders involved in consultations~\cite{bunea2015sharing}, while unions, associations, and other representative bodies can help align interests across similar organizations and coordinate their positioning.\\
In the EU regulatory context, platform firms deploy distinct influence strategies toward different audiences (lobbying for policymakers, coalition building for industry peers, and public relations for broader publics) illustrating how audience structure shapes positioning~\cite{gorwa2024platform}. Similarly, EU agency consultations can attract different stakeholder profiles depending on the policy stage: early-stage consultations draw more nonbusiness interests and experts, while late-stage implementation consultations are dominated by regulated industries~\cite{beyers2020feeds}.
Public emphasis on AI risks can itself become a form of positioning, as businesses facing competitive pressure may amplify public concern to signal industry leadership on safety,  establishing simultaneously the importance of the issue and their legitimacy as actors capable of managing it. Recently, Anthropic has provided a salient example of this, as its public communication foregrounds potentially catastrophic AI risks while presenting its own safety frameworks as models for responsible frontier development~\cite{de2026safety}, and urging governments to act promptly on AI regulation\footnote{See: \url{https://www.anthropic.com/news/the-case-for-targeted-regulation}}. Conversely, academics, experts, and NGOs, whose credibility depends on perceived authenticity, may face penalties for exaggeration and therefore be more cautious by showing restraint in public communication~\cite{brysse2013climate}.
Multiple audiences thus provides a mechanism through which publicly expressed concerns can diverge from positions elicited in surveys, as stakeholders calibrate their public communication to multiple audience-related incentive structures.

\paragraph{Reputational Risks}

The costs and benefits of expressing political attitudes and preferences differently across (public vs private) communicative settings~\cite{dur2013gaining}---and of potentially being exposed--- may vary systematically across stakeholder types and national contexts; as social norms and public observability can alter the incentives to misrepresent attitudes and preferences~\cite{valentim2024political}. For companies and business associations, appearing insufficiently attentive to AI-related risks may expose them to consumer distrust~\cite{rohit2026dark} and investor scrutiny over AI governance\footnote{See: \url{https://corpgov.law.harvard.edu/2025/12/22/a-look-at-ai-related-shareholder-proposals-at-u-s-companies-2022-2025}}. Alphabet Inc., for example, has faced investor pressure to strengthen board-level oversight of responsible AI development, with shareholders explicitly framing inadequate AI-risk governance as a source of reputational and business risk \footnote{See: \url{https://www.sec.gov/Archives/edgar/data/1652044/000121465926005268/o429268px14a6g.htm}}.
Recent work on regulatory agencies shows that organizations facing reputational threats deploy communicative responses calibrated to threat severity and audience expectations~\cite{bach2022regulatory}. Organizations in controversial sectors further employ symbolic risk management strategies~\cite{godfrey2009relationship}, such as stressing CSR and socially responsible values to direct attention away from controversial conduct. Overstating concerns for societal issues can also provide an ``insurance-like'' buffer against future adverse events. This can be seen as a form of ``defensive amplification'' consistent with proactive impression management strategies~\cite{mcdonnell2013keeping}. Public authorities, by contrast, may have incentives for moderation in public communication settings, to avoid reputational risks \cite{bach2022regulatory,gilad2015organizational} related to being perceived as alarmist, excessively interventionist, anti-innovation, or anti-business. This asymmetry suggests positive gaps for businesses and negative gaps for regulators,  when publicly communicating about AI risks, explainability, and related concerns.

\paragraph{Institutional Access Channels}

Organizations with established informal ties to policymakers can rely on private channels, such as industry working groups and expert committees, to communicate their positions, reducing the need for strong public advocacy~\cite{beyers2004voice,dur2013gaining}. Access, however, is both unequally distributed and context dependent. Evidence from EU platform lobbying illustrates this stratification: large firms obtain repeated high-level meetings with Commissioners, whereas smaller associations are more often directed to lower-ranking officials or denied access altogether~\cite{gorwa2024platform}. Organizations with less direct access may therefore have stronger incentives to rely on, and amplify their positions through, public-facing communication. This perspective also helps explain why business associations, despite having established access to policymakers, may amplify their publicly expressed concerns in settings such as in the context of the AI Act. Moreover, in the emerging AI regulatory domain, where established access may not always equally translate into effective engagement~\cite{sullivan2025nature}, business associations may complement inside channels with stronger public-facing expressions of concern. Business actors and associations with limited access may rely more heavily on public advocacy, while even comparatively well-connected actors may amplify concerns when established channels are uncertain or insufficiently effective.

\section{Material and methods}

\subsection{Data}

We use public consultations on the proposed European AI Regulation 2020, released via the Commission's Better Regulation Portal\footnote{\href{https://ec.europa.eu/info/law/better-regulation/have-your-say/initiatives/12270-White-Paper-on-Artificial-Intelligence-a-European-Approach/public-consultation_en}{European Commission, public consultation on the White Paper on Artificial Intelligence}.} and web-crawled from there. The dataset contains $N = 857$ submissions from 20-02-2020 -- 27-04-2021 from 3 rounds. Of these, $N = 348$ ($40.6\%$) contain both free-text responses and structured survey items measuring safety concern, rights concern, and perceived explainability importance on a 0--5 scale. These 348 documents form our validation sample.

\subsection{LLM annotation}

We prompted Qwen3.5-397b-A17b to annotate each document on three continuous scales (0--5): \textit{safety\_concern}, \textit{rights\_concern}, and \textit{explainability\_trust}. The prompt (Appendix~\ref{app:prompt}) required strict JSON output, included scale anchors, and specified binary flags for regulatory and biometric mentions. To assess inter-run LLM annotation alignment, we ran five independent annotation passes. Inter-run agreement was exceptionally high: ICC(C,1) ranged from $0.994$ to $0.996$, and ICC(C,k) exceeded $0.999$ for all constructs, indicating that the LLM was internally consistent---but consistency is not validity.\footnote{Mean vs. median aggregation across 5 LLM runs produces near-identical scores (difference \(< 0.01\) on 0--5 scale), confirming symmetric annotation distributions.}

\subsection{Validation framework}

Before using LLM annotations as proxies for survey constructs, we examine four diagnostic questions that help determine what the annotation is actually measuring. 

\paragraph{Construct correspondence}

Do the text-derived measure and the validation instrument refer to the same underlying quantity? For example, a prompt asking whether a text mentions AI safety does not necessarily measure how strongly the author reports being concerned about AI safety. This mismatch is particularly likely when the text is produced within institutional cultures---regulatory, bureaucratic, or professional---that shape communicative conventions. An LLM trained on general web text may not share the genre conventions of consultation submissions, creating a gap between the model's interpretive priors and the institutional context of the text.

\paragraph{Convergent and discriminant validity}

Where an external instrument exists, we assess convergence using complementary measures of association and agreement (Pearson's $r$, Lin's CCC, Bland--Altman limits). We also examine whether the proxy correlates more strongly with its target construct than with related constructs~\cite{lin2026validity}.

\paragraph{Systematic divergence}

Agreement should not be assessed through a single correlation coefficient alone. We examine whether misalignment between text and reference measures exhibit systematic bias or group-specific patterns. Systematic divergence may indicate measurement problems, but it may also reveal meaningful structure in the relationship between public communication and the reference instrument.

\paragraph{Downstream robustness}

We examine whether theoretically expected relationships involving the reference measure persist across reasonable specifications and subgroups. This helps distinguish substantively informative measures from measures that generate associations through specification or measurement artifacts.

\subsection{Spatial analysis}

Moran's $I$ was computed using queen contiguity weights (shared land borders define neighbors) via custom implementation with row-standardization. Permutation-based inference ($B = 9{,}999$ randomizations) provides Monte Carlo two-sided $p$-values. Primary analysis includes $N = 23$ countries with estimable divergence scores and at least one contiguous neighbor.

\subsection{Double machine learning specification}

We estimate the association between high survey-reported safety concern and survey-reported explainability support using Double/Debiased Machine Learning~\cite{chernozhukov2018double}. Pre-treatment covariates include organizational characteristics, capabilities, and activity areas. We estimate the propensity score and outcome regression using LightGBM with 5-fold cross-fitting. Causal Forest DML uses $N_{\text{trees}} = 1000$ and $\min_{\text{leaf}} = 10$.

\section{Results}

\subsection{Reproducibility and validity of LLM annotations}
\label{sec:gap}

Table~\ref{tab:validity} reports the agreement between LLM-derived text scores and survey items in the validation sample ($N = 348$). The correlations are far below accepted thresholds for proxy validity. Safety concern shows zero cross-setting correlation ($r = 0.029, p = 0.59$), while rights and explainability reach only weak correlation levels ($r = 0.176, p = 0.001$ and $r = 0.097, p = 0.072$, respectively). Lin's CCC---which penalises both location and scale bias---is near zero across all constructs (range: $0.013$ to $0.080$), indicating that, despite prompt alignment with the survey constructs, the LLM-derived and survey-based measures do not capture those constructs equivalently. Intraclass correlations confirm this pattern: ICC(2,1) and ICC(3,1) values remain below $0.15$ for all constructs.\\

\begin{table}[!ht]
\centering
\setlength{\tabcolsep}{4pt}
\caption{Convergent validity of LLM proxies against survey items ($N = 348$). Bonferroni-adjusted significance: $^{***}p < 0.0167$.}
\label{tab:validity}
\begin{tabular*}{\textwidth}{@{\extracolsep{\fill}}lccccc@{\extracolsep{\fill}}}
\toprule
\textbf{Construct} & \textbf{Pearson $r$} & \textbf{Lin's CCC} & \textbf{ICC(2,1)} & \textbf{ICC(3,1)} & \textbf{Cohen's $d$} \\
\midrule
Safety concern & $0.029$ & $0.013$ & $0.013$ & $0.027$ & $1.05$ \\
Rights concern & $0.176^{***}$ & $0.080$ & $0.080$ & $0.146$ & $0.98$ \\
Explainability & $0.097$ & $0.041$ & $0.041$ & $0.093$ & $1.18$ \\
\bottomrule
\end{tabular*}
\end{table}

Bland-Altman analysis reveals substantial systematic bias across all constructs, with Cohen's $d$ indicating standardized mean differences close to or exceeding one standard deviation for safety concern ($d = 1.05$), rights concern ($d = 0.98$), and explainability ($d = 1.18$). These effect sizes indicate that the bias is not merely a conservative threshold shift: it is large enough to substantially distort any estimates that treat LLM scores as valid proxies. The limits of agreement span approximately $\pm 4$ scale points, encompassing nearly the entire 0--5 measurement range.

Table~\ref{tab:cross}, which shows cross-construct correlations, provides further evidence of limited construct discrimination. The LLM's \textit{safety\_rights} score correlates almost as strongly with the survey's \textit{explain} item ($r = 0.162$) as with its own target ($r = 0.177$). This weak differentiation across constructs suggests that the LLM-derived scores may capture a broader dimension of expressed issue salience or concern, rather than cleanly distinguishing among the specific constructs targeted by the survey.

\begin{table}[!ht]
\centering
\setlength{\tabcolsep}{6pt}
\caption{Cross-construct correlations (LLM rows vs.\ survey columns, $N = 348$).}
\label{tab:cross}
\begin{tabular}{lccc}
\toprule
\textbf{LLM Proxy} & \textbf{Srvy Safety} & \textbf{Srvy Rights} & \textbf{Srvy Explain} \\
\midrule
LLM safety & $0.029$ & $-0.016$ & $0.011$ \\
LLM rights & $0.131$ & $0.177$ & $0.162$ \\
LLM explain & $0.072$ & $0.067$ & $0.096$ \\
\bottomrule
\end{tabular}
\end{table}
This weak differentiation across constructs may reflect differences in the evidentiary basis of the two measures, as they operationalize the same nominal construct differently. the LLM approach seeks to infer concern from its public textual expression, whereas the survey elicits the reported intensity of that concern through a standardized instrument. 
Our LLM annotation prompt asked, ``Does the text talk about safety risks?'' Placing the focus on the \emph{detection} and \emph{salience} of expressed concern in public text, whereas the survey asked: ``To what extent are you concerned about safety risks?'' seeking to directly elicit the \emph{intensity} of that concern. 
This shows, on the one hand, the inherent difficulty of framing an annotation prompt that inspects textual expression in exactly the same terms as a survey item that directly elicits reported intensity and, on the other, a broader measurement challenge: closely aligned constructs can still yield non-equivalent measures when they are elicited through different instruments and communicative settings~\cite{baden2022three,schwarz1999self}.
In the public communication setting, nearly every stakeholder who mentions safety in his  consultation submission also expresses concern; whereas variation in survey responses primarily reflects \emph{how strongly} stakeholders report that concern rather than simply \emph{whether} safety is mentioned or how salient it is in the text. As a result, the LLM-derived measure showed limited discrimination among higher levels of concern, compressing the upper end of the scale.

These differences help explain why text- and survey-based scores need not coincide, but they do not imply that the resulting misalignment is  random. 

This leads to a second empirical question:  whether this cross-setting divergence is itself systematically structured. We therefore construct a descriptive \emph{text--survey divergence  index} to examine this possibility. We emphasize that this index should not be interpreted as a direct measure of strategic intent or the extent to which a stakeholder ``amplifies'' their concern in public communication settings. Given the limited convergent validity between the LLM-derived and survey-based measures, cross-setting divergence may combine several sources of variation, including rhetorical emphasis, topic/issue selection, stakeholder type, country-level sociocultural factors, and cross-setting differences in public text vs. survey-based elicitation.

For each stakeholder $i$, we first standardize the component measures within the linked sample. The survey composite is then calculated as the unit-weighted mean of the standardized survey items on safety, rights, and explainability, while the corresponding text composite is calculated as the unit-weighted mean of the standardized LLM annotations of the same nominal domains. We then define

\begin{equation}
G_i =
\mathrm{TextScore}_i -
\mathrm{SurveyScore}_i .
\end{equation}

Positive values therefore indicate that the LLM-derived text score exceeds the stakeholder's standardized survey score, whereas negative values indicate the reverse. This distinction is important. If the text- and survey-based measures captured the same construct and differed only through classical measurement error, $G_i$ would primarily reflect random measurement noise. Our validation results provide evidence against this interpretation. Instead, the LLM-derived scores appear to capture features of consultation texts that are only partially aligned with the corresponding survey constructs.
We therefore examine whether $G_i$ varies systematically across stakeholder types. Such heterogeneity would suggest that the text–survey divergence is structured by organizational or institutional context rather than being purely idiosyncratic.

\subsection{Heterogeneity across stakeholder types}

\begin{figure}[!ht]
\centering
\includegraphics[width=0.7\linewidth]{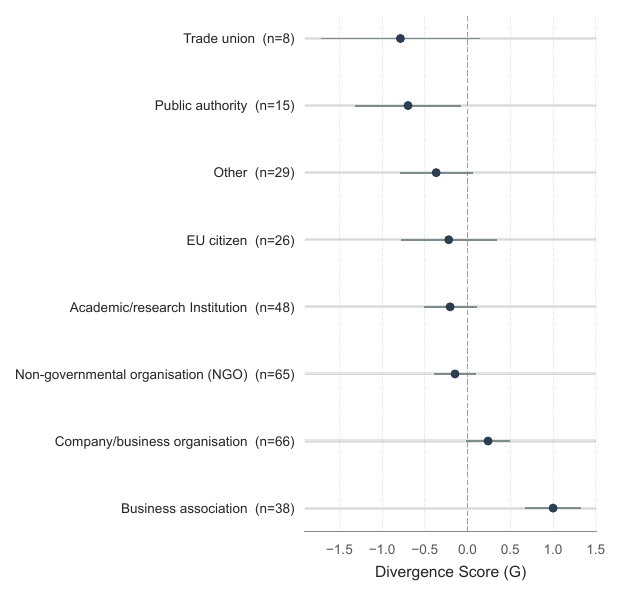}
\caption{Divergence score across stakeholder types. Points show mean differences between public-rhetoric and survey-reported concern composites (each constructed from standardized items), with horizontal bars denoting 95\% confidence intervals. Stakeholder types are ordered by the estimated divergence. Differences among stakeholder types were assessed using Welch's ANOVA followed by Games--Howell post hoc comparisons.}
\label{fig:gap}
\end{figure}

Figure~\ref{fig:gap} shows mean text---survey divergence by stakeholder type. The results reveal pronounced heterogeneity in how publicly expressed concerns relate to survey-reported concerns. Business associations show the largest positive divergence ($\bar{g} = +1.00$, 95\% CI $[0.68, 1.33]$, $N = 38$),indicating substantially greater emphasis on the targeted concerns in consultation texts than in their survey responses. Public authorities, by contrast, exhibit a sizeable negative divergence ($\bar{g} = -0.70$, 95\% CI $[-1.32, -0.07]$, $N = 15$). Mean divergence is also negative among academics ($\bar{g} = -0.20$, 95\% CI $[-0.52, 0.11]$, $N = 48$), NGOs ($\bar{g} = -0.15$, 95\% CI $[-0.39, 0.09]$, $N = 65$), EU citizens ($\bar{g} = -0.22$, 95\% CI $[-0.78, 0.34]$, $N = 26$), and trade unions ($\bar{g} = -0.79$, 95\% CI $[-1.72, 0.15]$, $N = 8$), although these estimates are individually compatible with no mean divergence. Games--Howell post-hoc tests show that business associations differ significantly from most other groups, whereas no other pairwise differences remain significant after accounting for unequal variances and sample sizes. Consistent with these findings, Welch's ANOVA indicates substantial heterogeneity across stakeholder types ($F = 7.93$, $p = 6.0 \times 10^{-7}$). Together, these results show that text---survey divergence is systematically structured by stakeholder type rather than being distributed randomly.

\subsection{Geographic patterns of text--survey divergence }

Beyond stakeholder-type heterogeneity, we examined whether divergence scores exhibit geographic patterning across the countries represented in our sample. As anticipated, this analysis is motivated by the possibility that national-level sociocultural and institutional factors shape communication differently across public and private settings. In particular, cross-country differences in norms surrounding public advocacy, institutional engagement, and reputational considerations may shape how stakeholders express their concerns differently in publicly facing consultation submissions relative to (more private) survey responses. We computed Moran's $I$ for country-level divergence scores using queen contiguity weights (countries sharing land borders are neighbors). Statistical significance was assessed using $B = 9{,}999$ random permutations, comparing the observed Moran’s $I$ with its reference distribution under spatial randomness to obtain permutation-based $p$-values. Analyses include $N = 23$ countries with estimable divergence scores; island nations without contiguous neighbors are excluded from spatial lag computation but retained in descriptive statistics.

Results reveal positive spatial autocorrelation: Moran's $I = 0.347$ ($p = 0.036$; expected $I$ under null $= -0.045$). This indicates that neighboring countries tend to exhibit more similar divergence scores than expected under spatial randomness, suggesting that public communication of AI-related concerns and explainability preferences is shaped by geographically clustered sociocultural, institutional, and regulatory contexts, rather than being purely organization-specific.

To assess whether this spatial clustering translates into broader regional differences, we grouped countries into five European regions: Anglo-Saxon, Nordic, Continental European, Southern European, and Eastern European countries. Table~\ref{tab:regional} presents divergence scores by region.
Anglo-Saxon countries show modest public-facing amplification ($\bar{g} = +0.29$, $N = 65$), while Continental European countries cluster near neutrality ($\bar{g} = +0.01$, $N = 163$). Southern European countries exhibit more pronounced public-facing restraint ($\bar{g} = -0.49$, $N = 36$), whereas Nordic countries show slight amplification ($\bar{g} = +0.20$, $N = 24$). Eastern European countries also show restraint ($\bar{g} = -0.35$, $N = 13$). Given the smaller sample sizes for Southern, Nordic, and Eastern European countries, these differences should be interpreted cautiously. Overall, these patterns suggest meaningful regional variation in text--survey divergence, although identifying the causal mechanisms underlying these differences remains beyond the scope of the present analysis.

\begin{table}[!ht]
\centering

\caption{Divergence score by region.}
\label{tab:regional}
\begin{tabular*}{\textwidth}{@{\extracolsep{\fill}}lrrr@{\extracolsep{\fill}}}
\toprule
\textbf{Region} & \textbf{Countries} & \textbf{Total $N$} & \textbf{Mean Divergence (SD)} \\
\midrule
Anglo         & 4 &  65 & $+0.29$ (0.12) \\
Continental   & 7 & 163 & $+0.01$ (0.38) \\
Nordic        & 4 &  24 & $+0.20$ (0.41) \\
Southern      & 6 &  36 & $-0.49$ (0.39) \\
Eastern       & 5 &  13 & $-0.35$ (0.56) \\
\bottomrule
\end{tabular*}
\footnotesize
\textit{Note.} Divergence scores are standardized (mean = 0, SD = 1). 
East Asia (Japan, $N = 1$) excluded from summary. 
Pairwise regional differences are descriptively large but not statistically 
significant after Bonferroni correction ($p_{\text{adj}} > 0.05$ for all contrasts).
\end{table}

\subsection{Robustness: Safety-explainability}
\label{sec:association}

Given the systematic divergence between LLM- and survey-based measures, we use the survey measures as the reference for downstream estimation. We estimate the association between high survey-reported safety concern (\texttt{concern\_safety} $\geq 4$) and survey-reported explainability support (\texttt{concern\_explain}), using Double/Debiased Machine Learning with pre-treatment covariates only (organizational characteristics, capabilities, and activity areas).

A Causal Forest DML estimator ($N_{\text{trees}} = 1000$, $\min_{\text{leaf}} = 10$) yields a coefficient of $1.39$ (95\% CI $[0.85, 1.93]$), indicating that stakeholders with high survey-reported safety concern show substantially greater support for explainability mandates. This association is robust across alternative specifications: Linear DML produces a similar estimate (coefficient $= 1.01$, 95\% CI $[0.65, 1.36]$), and permutation testing provides further evidence against the null ($p < 0.001$).

\begin{figure}[!ht]
\centering
\includegraphics[width=0.7\linewidth]{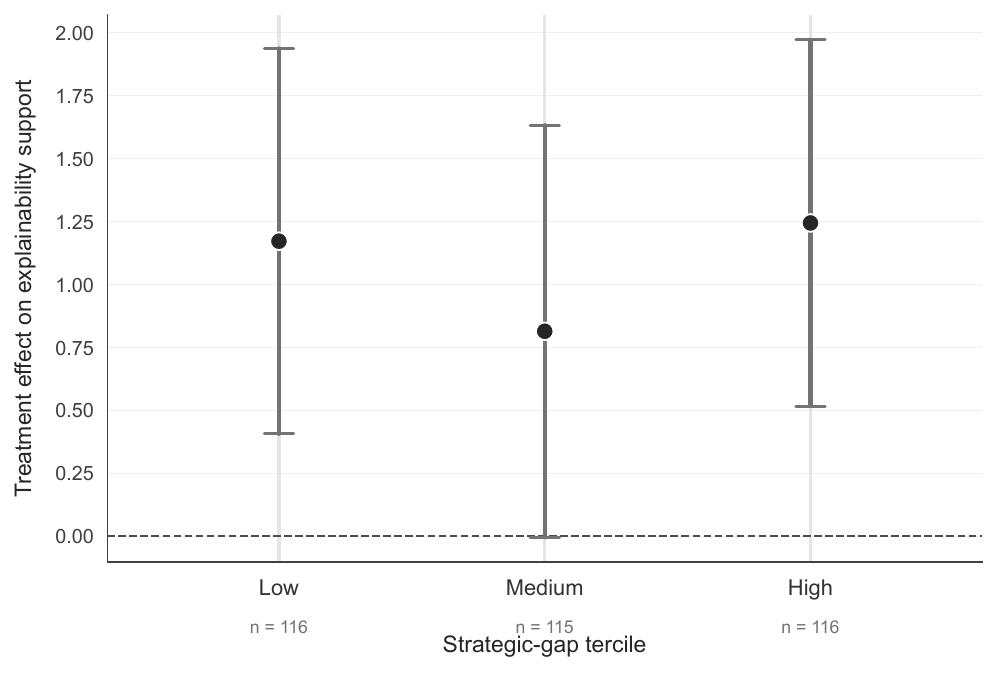}
\caption{Associations by divergence score tercile. Points indicate the estimated coefficient of high safety concern on support for explainability within terciles of the divergence score, with 95\% confidence intervals. The divergence score is defined as the standardized difference between public and survey assessments; terciles are ordered from low (more restrained) to high divergence. The horizontal dashed line indicates zero effect. Estimates were obtained using causal forest double machine learning within each tercile.}
\label{fig:cate}
\end{figure}

We then examine whether this association varies with the cross-setting divergence, by partitioning the sample into terciles and estimating coefficients within each tercile. Table~\ref{tab:cate_gap} and Figure~\ref{fig:cate} present the results. The low-divergence (most restrained) tercile shows coefficient $= 1.17$, 95\% CI $[0.41, 1.94]$. The medium tercile shows coefficient $= 0.82$, 95\% CI $[-0.004, 1.63]$ (marginally significant). Notably, the high-divergence tercile shows the \emph{strongest} effect: coefficient $= 1.25$, 95\% CI $[0.52, 1.97]$.

\begin{table}[!ht]
\centering
\footnotesize
\setlength{\tabcolsep}{2pt}
\caption{Coefficient of survey-reported safety concern on survey-reported explainability support, partitioned by divergence score tercile ($N = 347$).}
\label{tab:cate_gap}
\begin{tabular}{lcccc}
\toprule
\textbf{Divergence Tercile} & \textbf{Mean Divergence} & \textbf{Coefficient} & \textbf{95\% CI} & \textbf{$N$} \\
\midrule
Low divergence   & $-1.20$ & $1.172$ & $[0.407, 1.938]$ & $116$ \\
Medium divergence & $-0.03$ & $0.815$ & $[-0.004, 1.633]$ & $115$ \\
High divergence    & $+1.23$ & $1.245$ & $[0.518, 1.972]$ & $116$ \\
\bottomrule
\end{tabular}
\end{table}

This pattern suggests that the cross-setting divergence score does not simply attenuate associations through measurement error. Instead, stakeholders exhibiting greater public-facing amplification still show a strong association between survey-reported safety concern and support explainability: ---their public expression of concern is amplified, while their survey-reported regulatory positions remain coherently structured.\\

Our validation approach combined four complementary assessments. Convergent validity examines whether text-derived measures covary with survey measures targeting the same nominal construct (Pearson $r$).
Discriminant validity assesses whether each text-derived measure aligns more strongly with its intended survey construct than with other constructs.
Agreement analysis (Bland--Altman limits,  Lin's CCC, Cohen's $d$) evaluates whether the two public communication vs survey channels yield comparable scores, detecting systematic bias that correlation alone misses. Finally, downstream robustness checks examine whether theoretically expected relationships persist across specifications and subgroups. \\
The LLM-based text measures show limited convergent and discriminant validity and substantial disagreement with the survey measures despite exceptional inter-run agreement. At the same time, theoretically expected downstream relationships remained robust. Taken together, these findings show that highly reproducible text annotations can capture systematically structured information without being equivalent to survey-based measures: Even when targeting the same nominal constructs, surveys and public-texts may capture different context-dependent manifestations of them, such that limited convergence reflects not only measurement differences but also systematic variation in how concerns are expressed across communicative settings.

\section{Discussion}

The central finding of this study is a dissociation between reproducibility and construct correspondence in LLM-based measurement. Five independent annotation runs produced nearly identical scores, with intraclass correlations above $0.994$, yet the resulting text-derived measures showed little agreement with survey measures intended to capture the same nominal constructs. High agreement across independent LLM annotation therefore did not imply construct equivalence across measurement settings.
One source of cross-setting divergence lies in how the two measurement procedures operationalize the targeted constructs. The annotation task asks the LLM to infer safety, rights, and explainability concerns from their expression and salience in public consultation text, whereas the survey elicits stakeholders' reported intensity of concern through closed-ended items. These are closely related but non-equivalent measurement procedures, reflecting a broader distinction between LLM-based construct inference from public text, concern intensity elicited through closed-ended survey items, and the underlying constructs that both measurement approaches seek to operationalize~\cite{baden2022three}.
Importantly, neither measurement channel provides an unmediated observation of the underlying construct, as survey responses themselves are also shaped by question wording, response format, and elicitation context~\cite{schwarz1999self}.
A stakeholder can devote substantial textual space to a policy issue because the issue is institutionally salient, strategically relevant, legally consequential, or expected by the consultation format without necessarily reporting an equivalent (intensity of concern) score in a standardized survey.
This distinction changes how the LLM-derived text measure should be interpreted. Their limited correspondence with survey responses means that the annotation should not be be treated as a validated substitute for the corresponding survey measures. Yet limited convergence does not imply that the text measures are uninformative.
Consultation texts are institutionally situated communicative artifacts, and research on European interest-group politics shows that the articulation of policy positions varies systematically with organizational characteristics and inter-organizational ties~\cite{bunea2014explaining}, that the supply of policy-relevant information depends on organizational characteristics~\cite{kluver2012informational}, and that interest groups combine public-facing and access-oriented strategies in ways shaped by institutional opportunities and organizational type~\cite{beyers2004voice,dur2013gaining}.\\

Text--survey divergence may therefore contain systematic information about how related concerns are expressed across different communicative settings, by stakeholder type and country, that is not captured by either source in isolation.

Consistent with this interpretation, business associations exhibit the largest positive text--survey divergence in the linked sample, while public authorities and several non-business groups exhibit smaller or negative divergence. Such heterogeneity is consistent with prior evidence that organizational type structures modes of preference articulation~\cite{bunea2014explaining} and the balance between public-facing and access-oriented advocacy~\cite{dur2013gaining}. Organizational ties may further facilitate coordination and position alignment among stakeholders participating in consultations~\cite{bunea2015sharing}. These mechanisms are theoretically compatible with the observed pattern, but our design does not identify them directly. Differences in genre conventions, expertise, organizational roles, or interpretations of the survey and consultation tasks could generate similar patterns. We therefore interpret stakeholder-group differences as evidence of systematic institutional patterning rather than as direct evidence of strategic intent.\\

The geographic analysis provides weaker but complementary evidence. Country-level divergence  exhibits positive spatial autocorrelation in the primary specification, indicating that neighboring countries tend to display more similar patterns of text--survey divergence than expected under spatial randomness. This pattern is theoretically compatible with perspectives that understand political expression as embedded in socioculturally available repertoires of interpretation and action~\cite{swidler1986culture} and institutionally situated processes of political discourse. \\

Our findings speak to what LLMs measure when applied to institutionally situated communication. Public consultation texts combine substantive positions with organizational conventions, audience considerations, and the institutional conditions under which policy claims are articulated. LLMs applied to such texts therefore measure concerns and positions as manifested in communicative artifacts, rather than observing the underlying constructs independently of the contexts in which they are expressed. This distinction is central to longstanding concerns about measurement validity in computational text analysis~\cite{baden2022three}.  This distinction becomes more consequential when text-derived measures are used in causal analysis or policy evaluation~\cite{lin2025prompts}. When survey- and text-based measures diverge systematically, the interpretation of that divergence should depend on the inferential target. If the goal is to approximate a nominal construct as already operationalized by a reference instrument ---for example, the intensity of AI safety concern elicited through a closed-ended survey item--- divergence may be treated as measurement error relative to that benchmark. If, instead, the goal is to understand how the same nominal construct is expressed in public communication ---for example, how strongly AI risks or explainability are emphasized in consultation submissions--- the divergence itself becomes substantively informative.\\

More broadly, our work highlights three conditions under which high agreement across independent LLM annotation may coexist with limited construct validity: (i) when text annotations and surveys capture different manifestations of the same nominal construct; (ii) when public texts are produced under communicative and strategic incentives that differ from standardized survey elicitation; and (iii) when audience expectations or social desirability shape expression differently across measurement settings. 
Recognizing these conditions can help determine whether LLM-derived text measures capture meaningful variation in the nominal constructs researchers seek to measure, or whether their apparent validity instead reflects context-dependent expression, or other features of the measurement process.

\subsection*{Limitations}

Our analysis is limited to a single policy domain (EU AI regulation), a single consultation process, and a single LLM (Qwen3.5-397b-A17b). The extent to which the observed reproducibility--validity dissociation generalizes to other models, languages, policy domains, or genres of institutional communication is to be further investigated. The LLM prompt---'Does the text talk about safety risks?'---encodes an Anglo-American textual convention that treats explicit mention as evidence of concern. European regulatory rhetoric, by contrast, often embeds concern in procedural or precautionary language that an Anglo-calibrated model may miss or misweight. The resulting construct mismatch is therefore not merely a psychometric failure; it reflects a clash between the communicative conventions embedded in the model's training data and those of the institutional texts it annotates.
Survey items may themselves be affected by social desirability and ceiling effects, potentially reducing observed variation and attenuating associations among survey measures. Our analysis therefore demonstrates non-equivalence between two measurement procedures; it does not establish that the survey is an unbiased measure of an underlying latent construct. 
Importantly, our analysis does not establish an epistemic hierarchy between the two measurement approaches, nor does it imply that either should be treated a priori as privileged ground truth with respect to the nominal construct of interest. 

Because LLM-derived text scores emerge from a multi-stage measurement process, variation introduced at any stage ---from the communicative setting to the annotation protocol and scoring rules--- can alter their correspondence with survey-based measures. Modifying any one of these stages may therefore change the degree of measurement correspondence, but it does not resolve the more general distinction between the reproducibility of a measurement procedure and the validity of the inferences drawn from its output. The same logic applies to survey-based measures, which are likewise generated through specific elicitation processes. 

To assess sensitivity to prompt specification, we conducted a set of robustness checks using few-shot prompting and revised scale anchors. These checks were designed to probe the stability of the LLM-based measurement process and at no stage did we use survey correspondence as a criterion for prompt refinement. We deliberately avoided iterative prompt refinement against the survey responses because doing so would make benchmark alignment part of the optimization process. In settings where such optimization is pursued, specification choices should be separated across training, validation, and strictly held-out test data, with the final test sample insulated from model selection, prompt design, prompt/annotation choices, and scoring rules decisions.\\

While the aforementioned approach can reduce overfitting, it does not resolve the more fundamental problem of benchmark-oriented specification: With sufficiently large and representative training, validation, and held-out test sets drawn from the same measurement environment, a LLM-derived text measure can generalize well across all three while still being systematically shaped toward the benchmark used to define successful performance. Out-of-sample correspondence therefore establishes that benchmark alignment generalizes across observations, not that the resulting measure validly represents the nominal construct of interest. Transparency about prompt refinement, preregistration, independent validation and replication across samples or settings can constrain some researcher degrees of freedom, but none constitutes a definitive safeguard against LLM-based construct mismeasurement.
In light of these risks, we deliberately constrained the scope of our design and claims: within the measurement design evaluated here, high reproducibility provides evidence of procedural consistency, but is not sufficient to establish construct validity and alignment across measurement instruments.

\bibliographystyle{unsrtnat}
\bibliography{bibo, anthology, custom}
\appendix



\section*{Author contributions}
VB: Conceptualization, Data curation, Formal analysis, Investigation, Resources, Methodology, Software, Validation, Project Administration, Supervision, Funding Acquisition, Visualization, Writing – original draft, Writing – review \& editing. \\
CS: Conceptualization, Methodology, Writing – review \& editing.

\section*{Data availability}
The data and code underlying this article are available in a GitHub repository upon request to the authors.

\section*{Ethics Statement}
We use solely publicly available consultation data. No human subjects research requiring IRB approval was conducted. LLM annotations were performed programmatically without storing personal identifiers.

\section*{Usage of AI}

The study uses Qwen3.5-397B-A17B as the large language model for the text-annotation procedure described in the Methods. In addition, OpenAI ChatGPT was used during manuscript preparation to assist with proofreading, language editing and reframing. The authors independently evaluated and revised all AI-assisted text and take full responsibility for the content of the manuscript.

\section*{Competing interest}
The authors declare that they have no competing interests.

\newpage
\section{Supplementary Information}

\subsection{Initial LLM Annotation Prompt}
\label{app:prompt}

\begin{quote}
\small
\texttt{You are an expert annotator for European AI-policy consultation documents. \\
For each document you will be given a unique identifier (\texttt{doc\_id}) and the raw text. \\
Your job is to produce \textbf{exactly one JSON object} that follows the schema below:}

\vspace{0.5em}
\texttt{\{} \\
\texttt{  "doc\_id": "<the id you received>",} \\
\texttt{  "safety\_concern": <float 0--5>,} \\
\texttt{  "rights\_concern": <float 0--5>,} \\
\texttt{  "explainability\_trust": <float 0--5>,} \\
\texttt{  "topic\_distribution": \{ "<topic\_name>": <probability>, ... \},} \\
\texttt{  "flags": \{} \\
\texttt{    "mentions\_regulation": <true|false>,} \\
\texttt{    "mentions\_biometric": <true|false>} \\
\texttt{  \}} \\
\texttt{\}}

\vspace{0.5em}
\textit{Guidelines for the three scores (0 = no concern / no mention, 5 = very strong concern):}
\begin{itemize}
    \item \textbf{safety\_concern} -- Does the text talk about safety risks, accidents, or reliability of AI systems?
    \item \textbf{rights\_concern} -- Does the text discuss privacy, discrimination, fundamental-rights violations, or fairness?
    \item \textbf{explainability\_trust} -- Does the text emphasise the need for transparency, explainability, or trustworthiness of AI?
\end{itemize}

When a concept is not mentioned, output \textbf{0.0}. When unsure, round to nearest half-point (e.g., 2.5).

For \textbf{topic\_distribution}, assign probability to up to three topics that best summarise the document. Probabilities must sum to $\leq 1$.

For \textbf{flags}, set to \textbf{true} only if exact word appears (case-insensitive):
\begin{itemize}
    \item \texttt{mentions\_regulation} -- ``regulation'', ``regulatory'', or ``regulatory framework''
    \item \texttt{mentions\_biometric} -- ``biometric'' or ``biometric data''
\end{itemize}

Return \textbf{only the JSON object} -- no surrounding explanation, no markdown, no back-ticks.
\end{quote}
\newpage
\subsection{Revised Prompt (Post-Diagnosis)}

After discovering the validity failure, we revised the prompt to ask for \emph{predicted survey responses} rather than \emph{topic detection}. The revised prompt included explicit scale anchors (1--2 = brief mention, 4 = policy demand, 5 = central theme) and few-shot examples from the validation sample. Convergent validity improved marginally but remained below acceptable thresholds ($r = 0.046$ to $0.163$), suggesting that the problem is deeper than prompt engineering alone.

\begin{lstlisting}[style=promptstyle,
caption={System prompt and few-shot examples used for LLM-based stakeholder response prediction},
label={lst:llm-prompt}]
SYSTEM_PROMPT = """You are coding stakeholder submissions to the European Commission's AI Act public consultation. Each stakeholder also filled out a structured survey. Your task is to PREDICT their survey responses from the free text they provided.

OUTPUT FORMAT: Return exactly one JSON object, no markdown, no backticks, no explanation.

{
  "doc_id": "<the id you received>",
  "safety_concern": <float 0-5>,
  "rights_concern": <float 0-5>,
  "explainability_trust": <float 0-5>,
  "topic_distribution": { "<topic_name>": <probability>, ... },
  "flags": {
    "mentions_regulation": <true|false>,
    "mentions_biometric": <true|false>
  }
}

SCALE CALIBRATION (critical - follow exactly):
The survey uses a 6-point scale: 0 = not at all, 1 = slightly, 2 = somewhat, 3 = moderately, 4 = very much, 5 = extremely.

Most stakeholders who choose to write about a topic express genuine concern; therefore the distribution is heavily right-skewed. In this corpus:
- A score of 0 means the concept is literally not mentioned.
- 1-2 means brief, passing mention without argumentation.
- 3 means moderate concern with some reasoning.
- 4 means strong concern, explicit policy demand, or detailed argumentation.
- 5 means the issue is a central, dominant theme of the submission.

CONSTRUCT DEFINITIONS (distinguish carefully):

1. safety_concern - "To what extent are you concerned about safety risks, accidents, or reliability of AI systems?"
   - Code 0 if safety is not mentioned.
   - Code 1-2 for brief references (e.g., "we need safe AI").
   - Code 3 for moderate discussion with examples.
   - Code 4 for strong demands (e.g., "mandatory safety testing", "risk of accidents").
   - Code 5 if safety is the primary focus and drives most of the argument.

2. rights_concern - "To what extent are you concerned about privacy, discrimination, fundamental-rights violations, or fairness?"
   - Same scale. Distinguish from safety: rights concern is about human/legal entitlements, not technical failure.
   - Brief mention of GDPR or non-discrimination = 2-3.
   - Detailed argument about bias, fundamental rights, or privacy = 4-5.

3. explainability_trust - "To what extent does the text emphasise the need for transparency, explainability, or trustworthiness of AI?"
   - Code 0 if not mentioned.
   - Code 1-2 for passing reference to transparency.
   - Code 3 for moderate emphasis.
   - Code 4 for explicit demands for explainability mandates or citizen rights to explanation.
   - Code 5 if trust/explainability is the central policy demand.

TOPIC DISTRIBUTION: Assign up to 3 topics with probabilities \leq 1 (need not sum to 1). Use concise hyphenated labels (e.g., "health-AI", "biometric-identification", "autonomous-vehicles").

FLAGS (case-insensitive exact word match):
- mentions_regulation: true if "regulation", "regulatory", or "regulatory framework" appears.
- mentions_biometric: true if "biometric" or "biometric data" appears.

RULES:
- Base judgment ONLY on the submitted text. Do not infer from organisation type or country.
- If a concept is absent, output 0.0.
- Round to nearest 0.5 if uncertain.
- The output must be parseable JSON and nothing else.
"""


FEW_SHOT_EXAMPLES = """
EXAMPLES:

Example 1:
Text: "We support the AI Act. Safety is important but not our main focus. We are more worried about transparency in automated decision-making."
Survey prediction: {"doc_id": "EX1", "safety_concern": 2.0, "rights_concern": 1.0, "explainability_trust": 4.0, "topic_distribution": {"automated-decision-making": 0.6, "AI-Act": 0.3}, "flags": {"mentions_regulation": true, "mentions_biometric": false}}

Example 2:
Text: "The proposed regulation must ensure that AI systems in healthcare and transport are safe. Recent accidents with autonomous vehicles show the risks. Furthermore, citizens have a right to understand how AI affects their lives. Explainability is not optional."
Survey prediction: {"doc_id": "EX2", "safety_concern": 4.5, "rights_concern": 2.0, "explainability_trust": 4.0, "topic_distribution": {"autonomous-vehicles": 0.4, "healthcare-AI": 0.3, "citizen-rights": 0.2}, "flags": {"mentions_regulation": true, "mentions_biometric": false}}

Example 3:
Text: "We oppose strict regulation. Innovation should not be stifled by bureaucratic oversight."
Survey prediction: {"doc_id": "EX3", "safety_concern": 0.0, "rights_concern": 0.0, "explainability_trust": 0.0, "topic_distribution": {"innovation-policy": 0.7}, "flags": {"mentions_regulation": true, "mentions_biometric": false}}
"""
\end{lstlisting}

\subsection{LLM Annotation Flags}

We further validated LLM annotations against internal consistency criteria. Of 4,285 annotations (5 repeated annotations of 857 documents, 295 (6.9\%) exhibited patterns indicative of strategic communication rather than measurement error: 120 documents (2.8\%) showed high rhetorical intensity combined with low specificity---a signature of strategic positioning. 

\begin{table}[htbp]
\centering
\setlength{\tabcolsep}{9pt}
\caption{Flags in LLM Annotations ($N = 4,285$)}
\label{tab:qc_flags}
\begin{tabular}{lrr}
\toprule
\textbf{Flag Type} & \textbf{$N$} & \textbf{\%} \\
\midrule
High specificity + omissions & 145 & 3.4 \\
High rhetoric + low specificity & 120 & 2.8 \\
Very high concern + low rhetoric & 30 & 0.7 \\
\midrule
Any flag & 295 & 6.9 \\
\bottomrule
\end{tabular}
\end{table}

\subsection{Additional Robustness Results}
\label{app:robustness}

\subsection{Propensity Score Diagnostics}

\begin{figure}[!ht]
\centering
\includegraphics[width=0.8\linewidth]{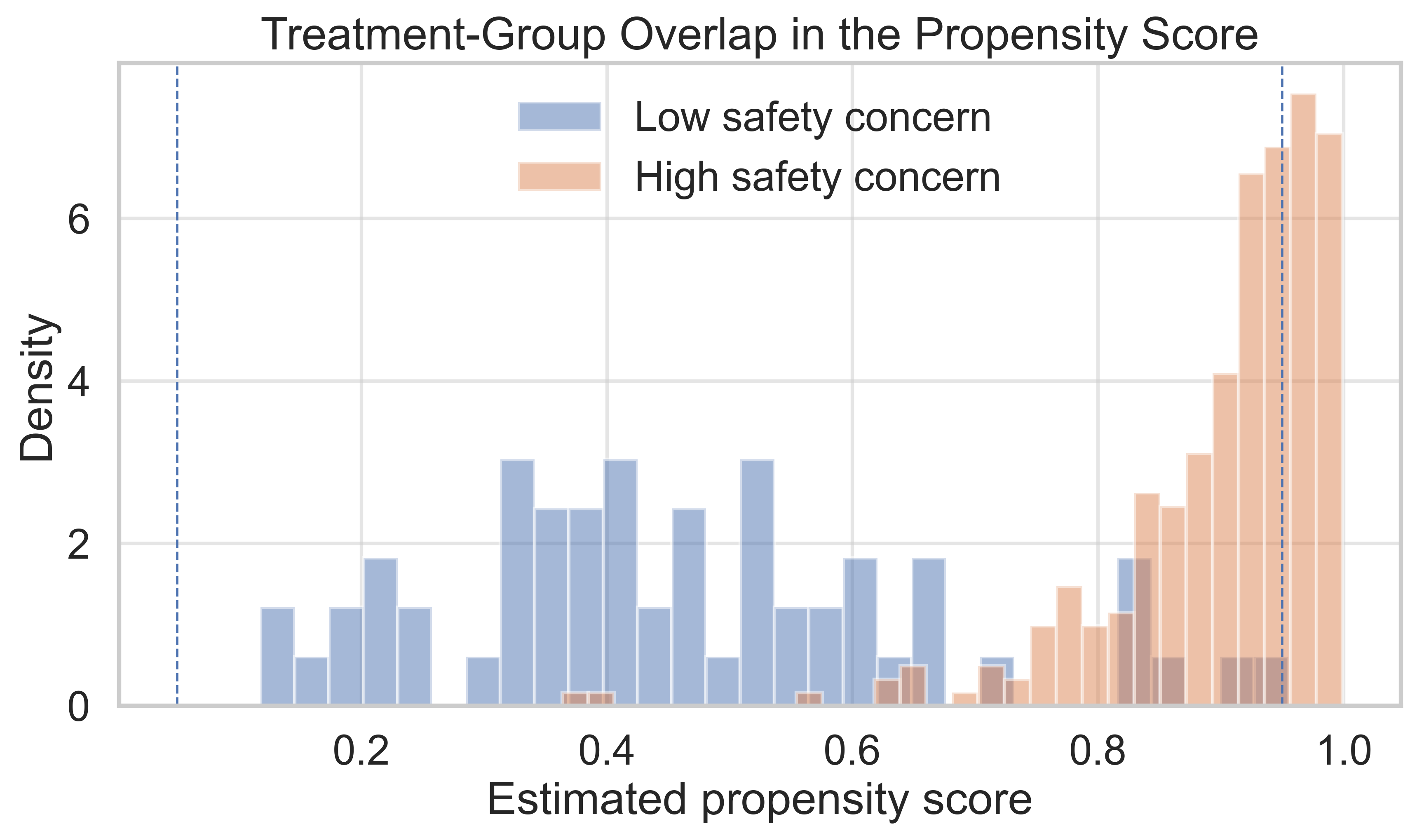}
\caption{Propensity score overlap between treated and control groups. The distribution shows adequate common support, with 70\% of observations falling within the 0.05--0.95 range (retained for trimmed analyses). Logistic regression propensity models achieved AUC $= 0.76$.}
\label{fig:prop_score}
\end{figure}

Propensity score overlap between treated and control groups was adequate, with 70\% of observations falling within the 0.05--0.95 common support region (see Fig.~\ref{fig:prop_score}). Logistic regression propensity models achieved AUC $= 0.76$, and LightGBM achieved AUC $= 0.66$.

\paragraph{Common support trimming.}
To assess whether effect estimates rely on extrapolation beyond the observed propensity score distribution, we re-estimated the coefficient after trimming observations with extreme propensity scores. Retaining observations with propensity scores between 0.05 and 0.95 (243 of 347 units, 70\%) yielded coefficient $= 0.58$ (95\% CI $[0.37, 0.79]$). Stricter trimming (235 units) produced coefficient $= 0.84$ (95\% CI $[0.64, 1.03]$). All trimmed estimates remain positive and statistically significant, though attenuated relative to the full sample (coefficient $= 1.01$, 95\% CI $[0.65, 1.36]$). This pattern indicates that while some extrapolation occurs at the propensity score boundaries, the core finding---that safety concern positively predicts explainability support---does not depend on extreme cases.

\subsection{Covariate Balance}

Standardized mean differences for all 843 covariates (including one-hot encoded categorical features) were reduced after inverse propensity weighting, with the largest initial imbalances showing substantial improvement. Inverse propensity weighting improved covariate balance: mean absolute SMD decreased from 0.18 to 0.06. Residual imbalance remained for 12 of 843 covariates (SMD \(> 0.2\) post-weighting); sensitivity analyses excluding these covariates yielded qualitatively similar results.
\FloatBarrier
\subsection{Inter-run LLM annotation agreement}

Five independent LLM annotation runs showed consistency: ICC(C,1) $= 0.995$ for safety concern, $0.996$ for rights concern, and $0.994$ for explainability trust. ICC(C,k) exceeded $0.999$ for all constructs. Leave-one-run-out analysis confirmed stability: correlations between full and reduced aggregates exceeded $r = 0.999$ for all constructs.

\subsection{Attrition Flow}

\begin{table}[htbp]
\setlength{\tabcolsep}{8pt}
\centering
\caption{Sample Attrition Flow}
\label{tab:attrition}
\begin{tabular}{lr}
\toprule
\textbf{Stage} & \textbf{$N$ (\%)} \\
\midrule
All consultation submissions & 857 (100.0\%) \\
Matched to LLM annotations & 857 (100.0\%) \\
Survey safety concern observed & 348 (40.6\%) \\
Survey explainability outcome observed & 348 (40.6\%) \\
Complete primary causal sample & 347 (40.5\%) \\
\bottomrule
\end{tabular}
\end{table}
\newpage
\FloatBarrier

\subsection{Explainability Support Across Safety Concern}

\begin{figure}[h!]
\centering
\includegraphics[width=0.7\linewidth]{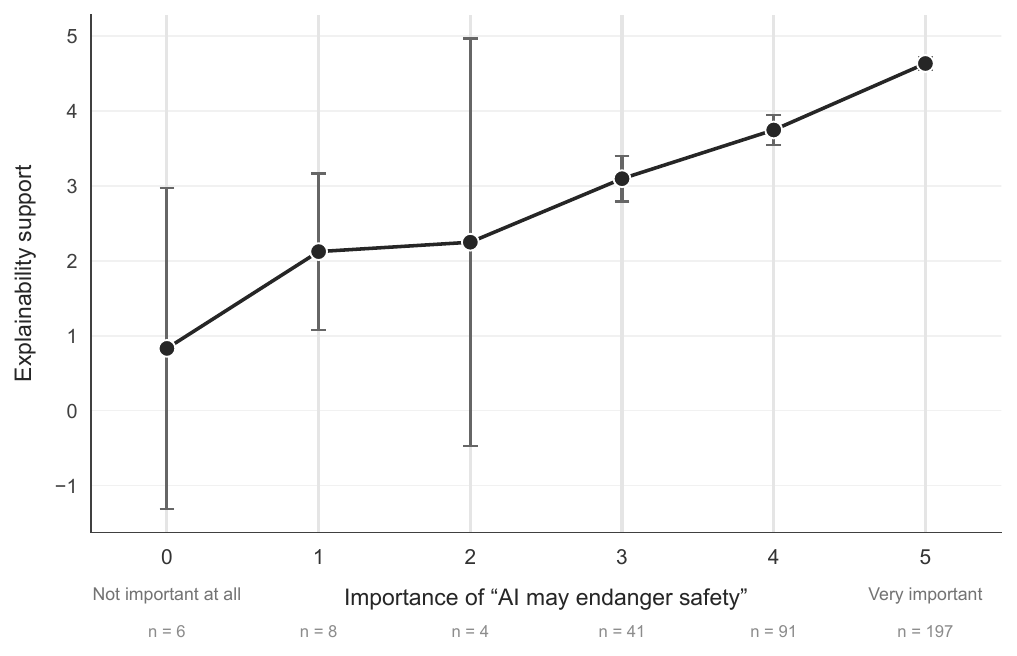}
\caption{Explainability support across levels of perceived safety concern. Points show mean explainability support for respondents reporting each level of importance of the statement "AI may endanger safety" (1 = not important at all; 5 = very important), with 95\% confidence intervals.}
\label{fig:explainabi}
\end{figure}

\FloatBarrier
\subsection{Supplementary Tables}
\label{app:tables}

\begin{table}[!htbp]
\centering
\caption{Robustness Analysis: Association of High Safety Concern with Explainability Support}
\label{tab:robustness_full}
\begin{tabular*}{\textwidth}{@{\extracolsep{\fill}}lrrrrr@{\extracolsep{\fill}}}
\toprule
\textbf{Specification} & \textbf{Effect} & \textbf{CI Lower} & \textbf{CI Upper} & \textbf{$N$} & \textbf{Significant} \\
\midrule
\multicolumn{6}{l}{\textit{Panel A: Primary Models}} \\
Linear DML & 1.006 & 0.651 & 1.361 & 347 & *** \\
Causal Forest DML & 1.392 & 0.853 & 1.930 & 347 & *** \\
\addlinespace
\multicolumn{6}{l}{\textit{Panel B: Threshold Variations (Safety Concern $\geq$ X)}} \\
Threshold $\geq$ 3.5 & 1.006 & 0.651 & 1.361 & 347 & *** \\
Threshold $\geq$ 4.0 & 1.006 & 0.651 & 1.361 & 347 & *** \\
Threshold $\geq$ 4.5 & 1.397 & 1.315 & 1.479 & 347 & *** \\
Threshold $\geq$ 5.0 & 1.397 & 1.315 & 1.479 & 347 & *** \\
\addlinespace
\multicolumn{6}{l}{\textit{Panel C: Common Support Trimming}} \\
Full sample & 1.006 & 0.651 & 1.361 & 347 & *** \\
0.05--0.95 propensity support & 0.579 & 0.370 & 0.788 & 243 & *** \\
Strict common support & 0.836 & 0.640 & 1.033 & 235 & *** \\
\addlinespace
\multicolumn{6}{l}{\textit{Panel D: Alternative Specifications}} \\
Continuous safety concern (per unit) & 1.477 & 1.430 & 1.523 & 347 & *** \\
Raw difference in means & 1.676 & 1.313 & 2.039 & 347 & *** \\
Permutation test $p$-value & \multicolumn{5}{l}{$< 0.001$} \\
\bottomrule
\end{tabular*}
\footnotesize
\textit{Note.} The threshold $\geq$ 3.0 specification produced an inadmissible estimate due to quasi-complete separation and is excluded. Adjusted OLS failed to converge because of near-collinearity between covariates and is omitted.
\end{table}

\begin{table}[htbp]
\centering
\caption{Agreement Statistics Between Survey Responses and LLM Annotations ($N = 348$) (Bonferroni-adj. $\alpha = 0.0167$)}
\label{tab:validation_full}
\scriptsize
\begin{tabular*}{\textwidth}{@{\extracolsep{\fill}}lrrrrrrr@{\extracolsep{\fill}}}
\toprule
\textbf{Construct} & \textbf{Pearson $r$} & \textbf{Spearman $\rho$} & \textbf{Lin's CCC} & \textbf{ICC(2,1)} & \textbf{ICC(3,1)} & \textbf{Cohen's $d$} & \textbf{$p$-adj} \\
\midrule
Safety & 0.029 & 0.015 & 0.013 & 0.013 & 0.027 & 1.05 & 1.0000 \\
Rights & 0.176 & 0.144 & 0.080 & 0.080 & 0.146 & 0.98 & 0.0031*** \\
Explainability & 0.097 & 0.066 & 0.041 & 0.041 & 0.093 & 1.18 & 0.2118 \\
\midrule
\multicolumn{8}{l}{\textit{Bland-Altman Bias Analysis}} \\
\midrule
Safety & \multicolumn{2}{c}{Bias = 1.98} & \multicolumn{2}{c}{LoA = [$-$1.71, 5.67]} & \multicolumn{3}{c}{$d$ = 1.05} \\
Rights & \multicolumn{2}{c}{Bias = 1.79} & \multicolumn{2}{c}{LoA = [$-$1.78, 5.36]} & \multicolumn{3}{c}{$d$ = 0.98} \\
Explainability & \multicolumn{2}{c}{Bias = 2.15} & \multicolumn{2}{c}{LoA = [$-$1.42, 5.72]} & \multicolumn{3}{c}{$d$ = 1.18} \\
\midrule
\multicolumn{8}{l}{\textit{Inter-Run Agreement (ICC across 5 LLM runs)}} \\
\midrule
Safety concern & \multicolumn{2}{c}{ICC(C,1) = 0.995} & \multicolumn{2}{c}{ICC(C,k) = 0.999} & \multicolumn{3}{c}{$p < 0.001$} \\
Rights concern & \multicolumn{2}{c}{ICC(C,1) = 0.996} & \multicolumn{2}{c}{ICC(C,k) = 0.999} & \multicolumn{3}{c}{$p < 0.001$} \\
Explainability trust & \multicolumn{2}{c}{ICC(C,1) = 0.994} & \multicolumn{2}{c}{ICC(C,k) = 0.999} & \multicolumn{3}{c}{$p < 0.001$} \\
\bottomrule
\end{tabular*}
\end{table}

\begin{table}[!htbp]
\centering
\caption{Divergence Score (Public Rhetoric $-$ Survey-Reported Concern) by Stakeholder Type}
\label{tab:gap_stakeholder_full}
\begin{tabular*}{\textwidth}{@{\extracolsep{\fill}}lrrrr@{\extracolsep{\fill}}}
\toprule
\textbf{Stakeholder Type} & \textbf{Mean} & \textbf{SD} & \textbf{N} & \textbf{95\% CI} \\
\midrule
Trade union & $-$0.786 & 1.118 & 8 & [$-$1.57, 0.00] \\
Public authority & $-$0.697 & 1.129 & 15 & [$-$1.28, $-$0.11] \\
Other & $-$0.368 & 1.122 & 29 & [$-$0.78, 0.05] \\
EU citizen & $-$0.221 & 1.390 & 26 & [$-$0.77, 0.33] \\
Academic/research Institution & $-$0.204 & 1.071 & 48 & [$-$0.51, 0.10] \\
Non-governmental organisation (NGO) & $-$0.148 & 0.974 & 65 & [$-$0.39, 0.09] \\
Company/business organisation & 0.239 & 1.034 & 66 & [$-$0.01, 0.49] \\
Business association & 1.001 & 0.987 & 38 & [0.68, 1.32] \\
\midrule
\multicolumn{5}{l}{\textit{ANOVA: $F(7, 287) = 7.93$, $p < 0.0001$}} \\
\midrule
\multicolumn{5}{l}{\textit{Pairwise Comparisons (Bonferroni-adjusted, 28 tests, $\alpha_{adj} = 0.0018$)}} \\
\midrule
\textbf{Comparison} & \textbf{$t$} & \textbf{$p_{raw}$} & \textbf{$p_{adj}$} & \textbf{Significant} \\
\midrule
Public authority vs Business association & $-$5.10 & 0.0000 & 0.0010 & *** \\
Other vs Business association & $-$5.21 & 0.0000 & 0.0001 & *** \\
EU citizen vs Business association & $-$3.86 & 0.0004 & 0.0107 & *** \\
Academic/research vs Business association & $-$5.42 & 0.0000 & 0.0000 & *** \\
NGO vs Business association & $-$5.72 & 0.0000 & 0.0000 & *** \\
Company/business vs Business association & $-$3.72 & 0.0000 & 0.0103 & *** \\
\bottomrule
\end{tabular*}
\end{table}
\subsection{Geographic Analysis Methods and Figures}
\label{app:geo-figures}


\begin{figure}[!ht]
\centering
\includegraphics[width=0.7\linewidth]{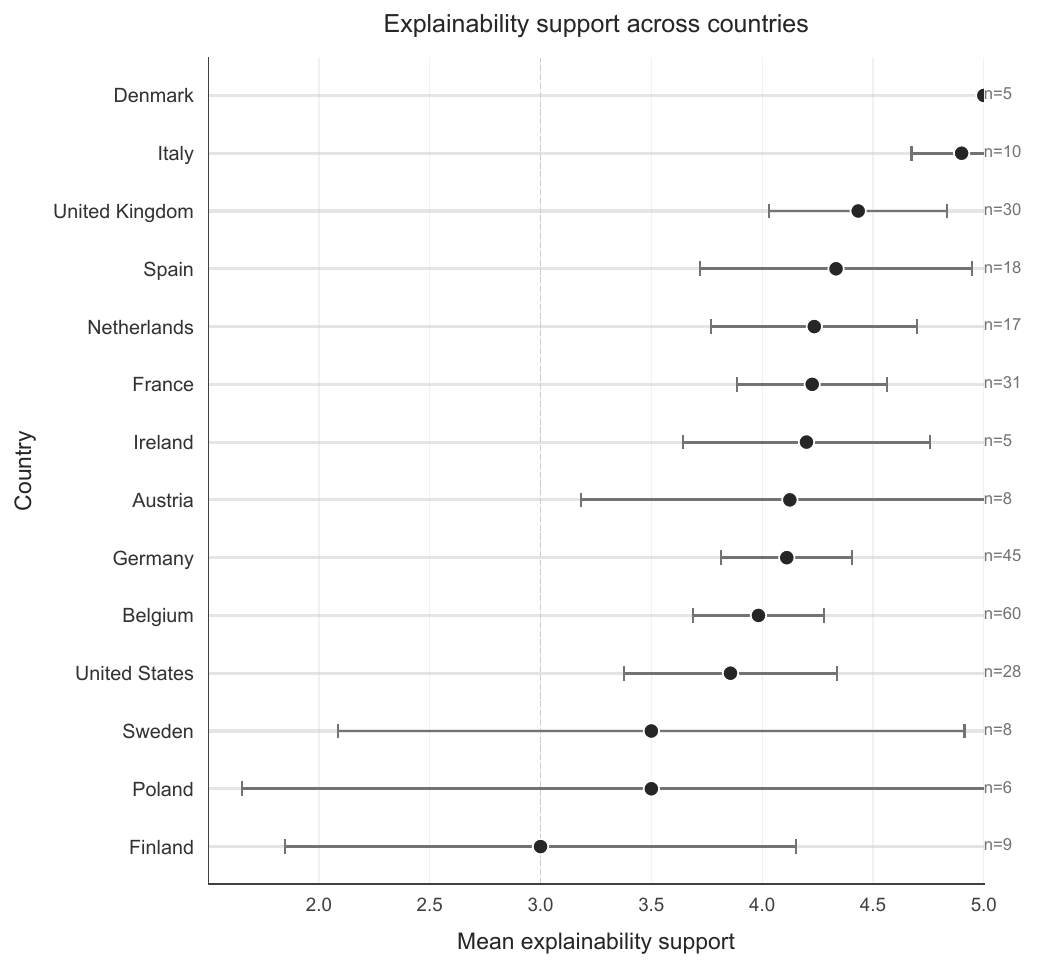}
\caption{Explainability support across countries. Points indicate the country-specific mean of explainability support, with 95\% confidence intervals based on the t distribution. Countries with fewer than five observations are omitted.}
\label{fig:support_by_country}
\end{figure}
\subsection{Country-Specific Associations}

Country-level estimation was feasible for only 3 of 27 countries due to limited within-country variation in treatment assignment. France shows a positive association ($\hat{\beta} = 1.80$, 95\% CI $[0.80, 2.79]$); Germany and Belgium show similar-magnitude but non-significant associations.

\subsubsection{Spatial Autocorrelation Computation}

Moran's $I$ was computed using queen contiguity weights (shared land borders define neighbors) via custom implementation with row-standardization. Permutation-based inference ($B = 9{,}999$ randomizations) provides exact two-sided $p$-values. Primary analysis includes $N = 23$ countries with estimable divergence scores and at least one contiguous neighbor. The positive Moran's $I = 0.347$ ($p = 0.036$) indicates significant spatial clustering rather than dispersion. Sensitivity analyses with alternative inclusion thresholds ($N \geq 5$ per country) yield qualitatively similar patterns ($I = 0.333$, $p = 0.060$).

\begin{figure}[!ht]
\centering
\includegraphics[width=1.4\linewidth]{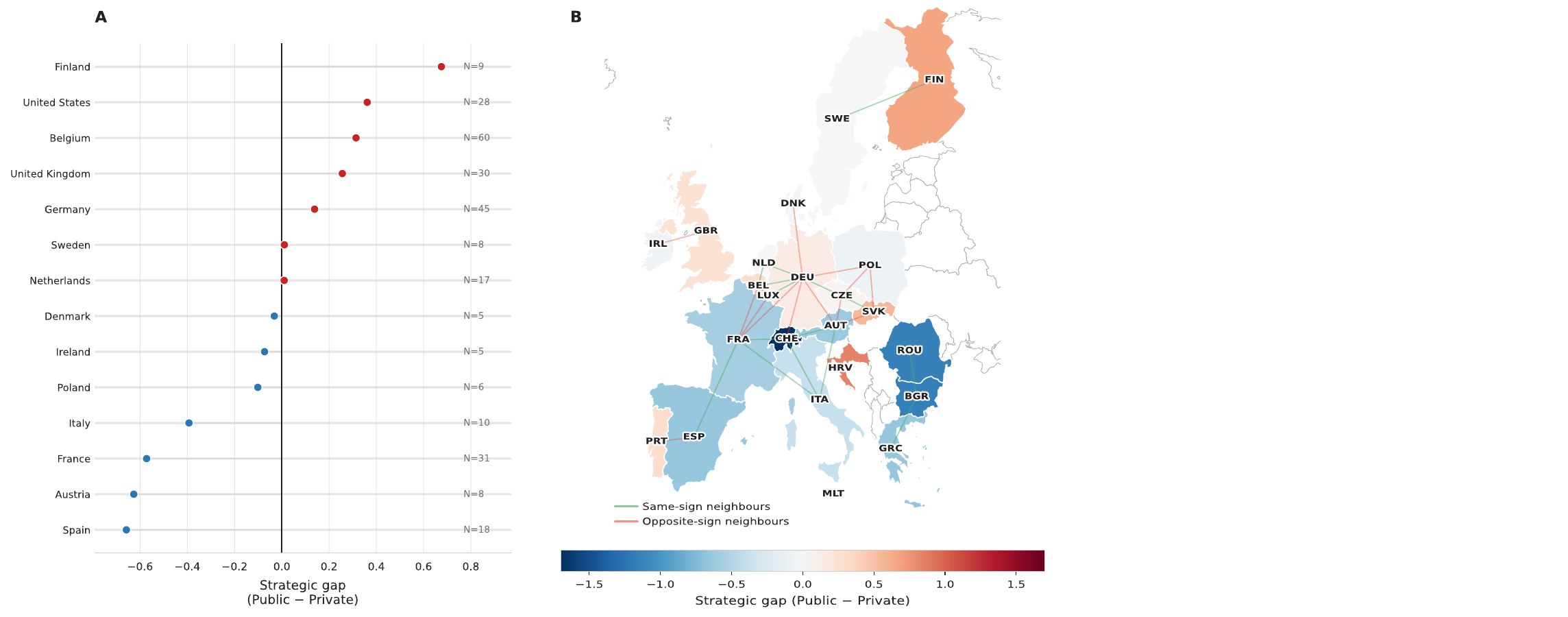}
\caption{
Spatial contiguity network and strategic gap clustering across Europe. Nodes represent $N = 23$ countries with divergence scores; edges denote shared land borders (queen contiguity weights). Node colors indicate strategic gap values (red = public rhetoric, blue = strategic restraint). Edge thickness reflects border length. Primary analysis (reported in main text) includes all countries with contiguous neighbors (Moran's $I = 0.347$, $p = 0.036$, $B = 9{,}999$ permutations). This network visualization applies an $N \geq 5$ per-country threshold for stability (Moran's $I = 0.333$, $p = 0.060$), shown here as a robustness check. Continental European countries (Germany, France, Belgium, Netherlands) form a moderate-inflation cluster; Southern European countries (Spain, Italy, Portugal) show restraint. Island nations (Ireland, UK) excluded from spatial lag computation due to lack of contiguous neighbors.
}
\label{fig:moran_network}
\end{figure}
\subsubsection{Regional Classification}

Countries grouped into cultural-institutional regions following standard comparative political economy typologies:
\begin{itemize}
    \item \textbf{Anglo}: United Kingdom, Ireland, United States, Canada
    \item \textbf{Nordic}: Sweden, Denmark, Finland, Norway
    \item \textbf{Continental}: Germany, France, Belgium, Netherlands, Austria, Switzerland, Luxembourg
    \item \textbf{Southern}: Italy, Spain, Portugal, Greece, Malta, Croatia
    \item \textbf{Eastern}: Poland, Czech Republic, Slovakia, Romania, Bulgaria
\end{itemize}

\subsubsection{Spatial Autocorrelation of Three Constructs}

\begin{figure}[!ht]
\centering
\includegraphics[width=0.85\linewidth]{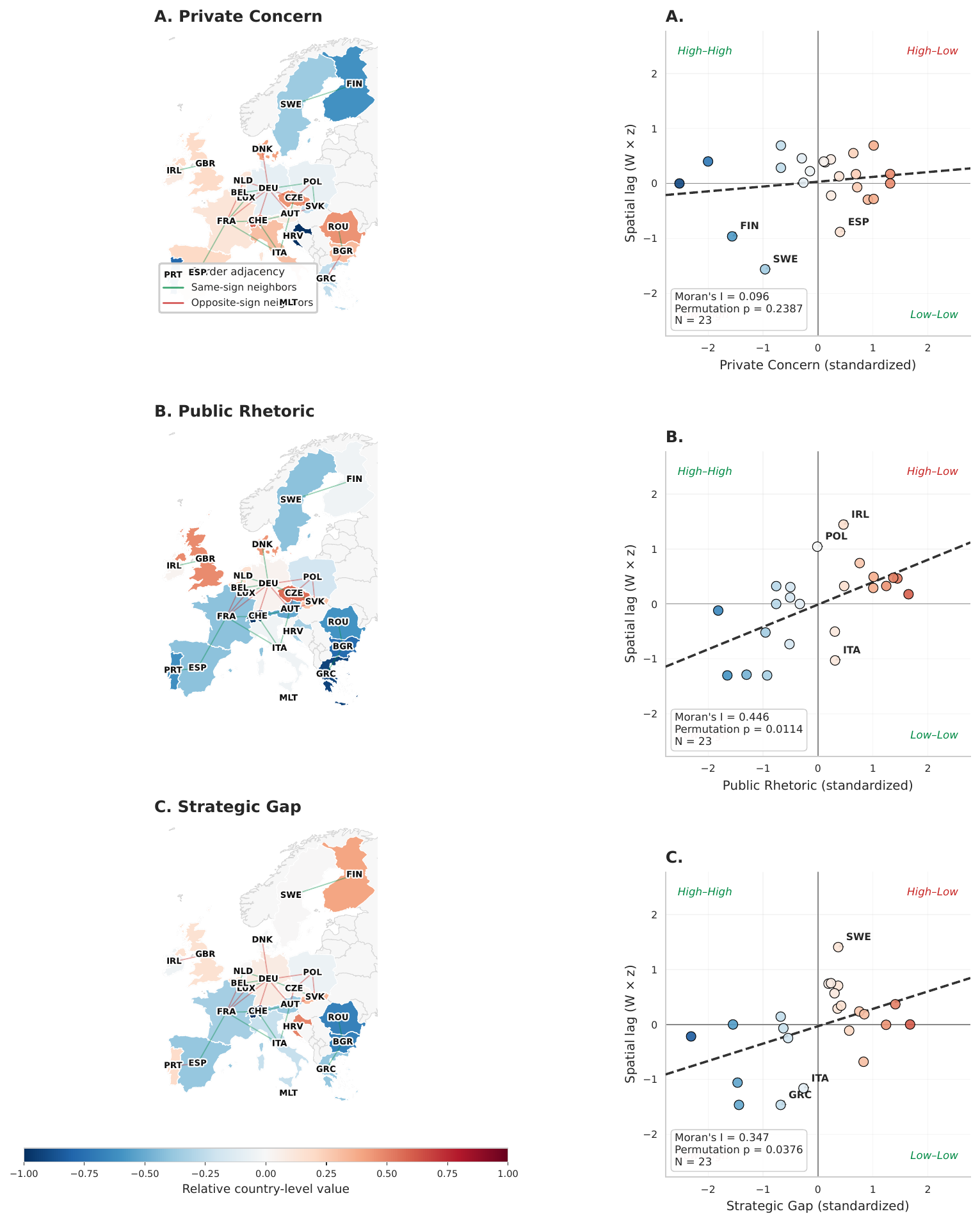}
\caption{Spatial autocorrelation of Survey-Reported Concern, Text-based Rhetoric, and Divergence Score. Moran scatterplots and permutation-based null distributions for $N = 23$ European countries ($B = 9999$ permutations). Survey-Reported Concern shows weak and nonsignificant spatial autocorrelation ($I = 0.096$, $p = 0.240$), whereas Public Rhetoric exhibits significant positive spatial autocorrelation ($I = 0.446$, $p = 0.009$). Divergence Score also shows positive spatial autocorrelation ($I = 0.347$, $p = 0.036$), but this association does not survive correction for multiple testing (FDR-adjusted $p = 0.054$; Bonferroni-adjusted $p = 0.108$). The pattern indicates stronger geographic clustering in public rhetoric than in survey-reported concern, with divergence showing an intermediate pattern.}
\label{fig:three_constructs_spatial}
\end{figure}

Appendix Figure~\ref{fig:three_constructs_spatial} shows that spatial clustering is specific to Public Rhetoric and Divergence Score, while Survey-Reported Concern exhibits no significant autocorrelation.

\end{document}